\documentclass[sn-mathphys-num]{sn-jnl}
\PassOptionsToPackage{table}{xcolor}
\usepackage{graphicx}%
\usepackage{multirow}%
\usepackage{amsmath,amssymb,amsfonts}%
\usepackage{amsthm}%
\usepackage{mathrsfs}%
\usepackage[title]{appendix}%
\usepackage{xcolor}%
\usepackage{textcomp}%
\usepackage{manyfoot}%
\usepackage{booktabs}%
\usepackage{algorithm}%
\usepackage{algorithmicx}%
\usepackage{algpseudocode}%
\usepackage{listings}%
\usepackage{subcaption}
\usepackage[table]{xcolor}
\usepackage{caption}
\usepackage{longtable}
\usepackage{array}
\usepackage{graphicx}
\usepackage{booktabs}
\graphicspath{./figures}

\theoremstyle{thmstyleone}%
\theoremstyle{thmstyletwo}%
\newtheorem{example}{Example}%

\theoremstyle{thmstylethree}%
\newtheorem{definition}{Definition}%
\newcommand{\RNum}[1]{\uppercase\expandafter{\romannumeral #1\relax}}

\begin{document}
\captionsetup[table]{labelfont=bf, font=normalsize,labelsep=period}

\title[MIMIC-\RNum{4}-Ext-22MCTS: A 22 Millions-Event Temporal Clinical Time-Series Dataset with Relative Timestamp for Risk Prediction]{MIMIC-\RNum{4}-Ext-22MCTS: A 22 Million-Event Temporal Clinical Time-Series Dataset for Risk Prediction}


\author*[1]{\fnm{Jing Wang} }\email{jing.wang20@nih.gov}
\author[2]{\fnm{Xing Niu} }
\author[3]{\fnm{Tong Zhang} }
\author[4]{\fnm{Jie Shen} }
\author[5]{\fnm{Juyong Kim} }
\author[1]{\fnm{Jeremy C. Weiss}}

\affil*[1]{\orgdiv{National Library of Medicine}, \orgaddress{ \city{Bethesda},  \state{MD}, \country{USA}}}
\affil[2]{\orgname{AWS AI Labs}, \orgaddress{\city{New York}, \state{NY}, \country{USA}}}
\affil[3]{\orgname{University of Illinois Urbana-Champaign}, \orgaddress{\city{Urbana}, \state{IL}, \country{USA}}}
\affil[4]{\orgname{Stevens Institute of Technology}, \orgaddress{\city{Hoboken},  \state{NJ}, \country{USA}}}
\affil[5]{\orgname{Carnegie Mellon University}, \orgaddress{\city{Pittsburgh}, \postcode{15213}, \state{PA}, \country{USA}}}

\abstract{Clinical risk prediction based on machine learning algorithms plays a vital role in modern healthcare. A crucial component in developing a reliable prediction model is collecting high-quality time series clinical events. In this work, we release such a dataset that consists of 22,588,586 Clinical Time Series events, which we term MIMIC-\RNum{4}-Ext-22MCTS. Our source data are discharge summaries selected from the well-known yet unstructured MIMIC-IV-Note \cite{Johnson2023pg}. We then extract clinical events as short text span from the discharge summaries, along with the timestamps of these events as temporal information. The general-purpose MIMIC-IV-Note pose specific challenges for our work: it turns out that the discharge summaries are too lengthy for typical natural language models to process, and the clinical events of interest often are not accompanied with explicit timestamps. Therefore, we propose a new framework that works as follows: 1) we break each discharge summary into manageably small text chunks; 2) we apply contextual BM25 and contextual semantic search to retrieve chunks that have a high potential of containing clinical events; and 3) we carefully design prompts to teach the recently released Llama-3.1-8B \cite{touvron2023llama} model to identify or infer temporal information of the chunks. We show that the obtained dataset is so informative and transparent that standard models fine-tuned on our dataset are achieving significant improvements in healthcare applications. In particular, the BERT model fine-tuned based on our dataset achieves 10\% improvement in accuracy on medical question answering task, and 3\% improvement in clinical trial matching task compared with the classic BERT. The GPT-2 model, fine-tuned on our dataset, produces more clinically reliable results for clinical questions. The dataset is available at \href{https://physionet.org/content/mimic-iv-ext-22mcts/1.0.0}{\textcolor{blue}{physionet.org/content/mimic-iv-ext-22mcts/1.0.0}}. The codebase is released at \href{https://github.com/JingWang-RU/MIMIC-IV-Ext-22MCTS-Temporal-Clinical-Time-Series-Dataset}{\textcolor{blue}{github.com/JingWang-RU/MIMIC-IV-Ext-22MCTS-Temporal-Clinical-Time-Series-Dataset}}.}

\keywords{Clinical event, Temporal information, Time series, MIMIC, Contextual BM25, Contextual semantic search, Natural language processing, Large language model, Question answering, Clinical trial}

\maketitle

\section{Introduction}\label{sec1}
	Clinical risk forecasting modeling with electronic health record (EHR) data is anticipated to drive personalized medicine, improve healthcare quality, and develop and update clinical practice guidelines. Developing machine learning models to forecast clinical risk and related timestamp typically requires extraction of clinical events and timestamps. The time series clinical events with temporal information help characterize patient trajectories and track the disease progression, sometimes even find the reasoning and consequence of disease. However,  acquiring such time series data is a labor-intensive process that requires not only significant effort but also a deep understanding of clinical and medical knowledge for accurate event time reasoning. 

In this work, we release a time series clinical events dataset with concrete temporal information. The dataset consists of 22,588,586 clinical events and related timestamps from 267,284 discharge summaries of  MIMIC-IV-Note \cite{Johnson2023pg}. We provide the definition of a clinical event as follows:
\begin{definition}
	\label{df}
	\textbf{Clinical event} is a free-text specification of an entity pertaining to or with the potential to affect the individual's health that can be temporally located.
\end{definition}

The clinical event usually is a short text span found in the discharge summary. The timestamp annotation represents a relative timestamp measured in hours, referenced from a major event. It can take negative values if the event occurs before the reference point. Each sample in the dataset is in the format as shown below: 
\begin{align}
[\text{TIME}] <\text{timestamp}> [\text{EVENT}] <\text{description}>
\end{align}
We present an example of clinical expert-annotated events and timestamps from a publicly available case report in Table \ref{tab:annote}. This is shown instead of our experimental dataset, which is based on the MIMIC database and subject to restricted access.

The source dataset is MIMIC-IV-Note, a collection of 331,794 de-identified discharge summaries from 145,915 patients admitted to the hospital and emergency department at the Beth Israel Deaconess Medical Center in Boston, MA, USA between 2008 and 2019. We select notes occurring within one year of a patient encounter. The notes need to have brief hospital courses section with more than 100 characters \cite{bhc2024}. It leads to a 267,284 discharge summaries in total. 

There are many challenges to annotate clinical events and related timestamp from the discharge summary, such as:
\begin{itemize}
	\item The discharge summaries in MIMIC-IV-Note have an average token length of 2,267 $\pm$ 914. The name entity recognition models, especially those based on transformers (e.g. BERT, BioBERT), have a maximum token limit, such as 512 tokens for BERT. If sliding windows or chunking techniques are used, it requires more computational resources and longer time to process. 

	\item The temporal information of related clinical events usually is not recorded in the original discharge summary. With the recent advancement of Large Language Models (LLMs), it has been recognized that LLM may serve as a useful computational tool for EHR annotation. However, applying LLM as a black box is risky: the generated text looks natural narrative but may not be the expected outcome, a phenomenon termed hallucination.
\end{itemize}
	
\begin{figure}[h]
		\centering
		\includegraphics[width=0.9\textwidth]{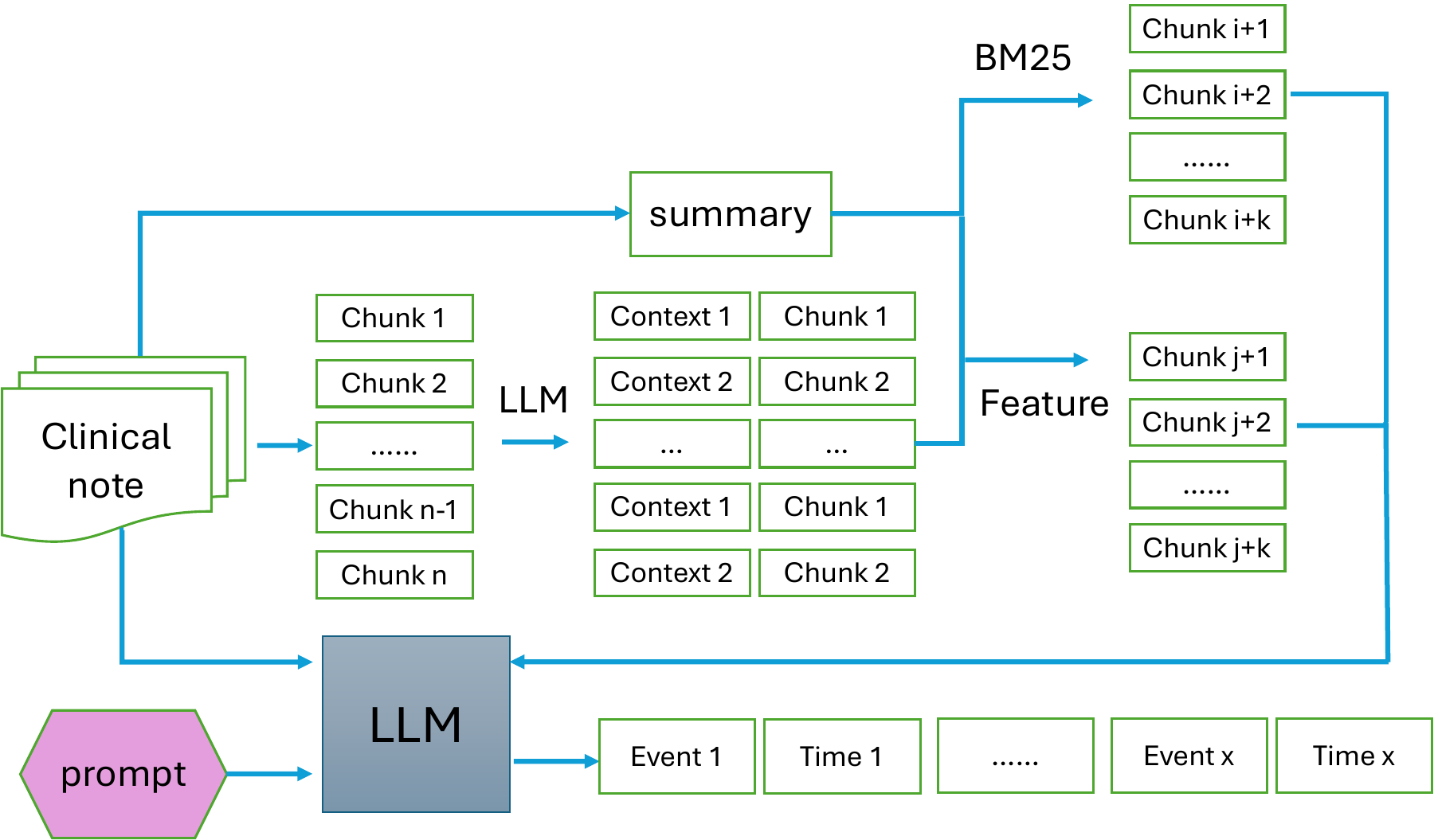} 
		\caption{The pipeline of end-to-end annotation framework.}
		\label{fig:rag}
	\end{figure}

	To solve the challenges, we propose an end-to-end framework that produces reliable annotation of time series clinical events and related timestamps for discharge summary based on retrieval and Large Language Models. Here is the brief description of the framework:
	\begin{itemize}
		\item Each summary is segmented into a series of chunks with maximal length 5 tokens, the previous 5 tokens and the following 5 tokens are treated as the context.
		\item The contextual BM25 retrieves top 100 chunks with high  probability to contain clinical events with brief hospital courses as query. 
		\item We utilize BGE-Large-en model \cite{bge_embedding} to learn the embedding of query (brief hospital courses) and contextual chunks, compute the correlation similarity score between query and chunks, and retrieve the chunks with correlation score higher than 0.75. The threshold of 0.75 is indicative of a high level of similarity.
		\item The chunks retrieved by contextual BM25 and semantic search are combined without duplications. 
		\item We design the prompt to teach the Large Language Model, Llama-3.1-8B \cite{touvron2023llama}\cite{grattafiori2024llama}  to identify chunks that contain clinical events, and estimate the relative timestamps of events.
	\end{itemize}
	The illustration of the framework is shown as Figure \ref{fig:rag}.
	To evaluate the qualify of the annotated dataset, we fine-tune a Transformer encoder, BERT \cite{devlin2019bert}, OpenAI released large Transformer decoder, GPT-2 \cite{radford2019language} on our datasets. Since the original BERT do not support temporal information, we include temporal embedding layer, concatenate the features with the embedding of clinical events, and feed them to a multi-layer perceptron to explore the causal relationship of the two events: reason/consequence/no correlation. We extend the GPT-2 model in a similar way. Then we compare the performance of vanilla BERT and GPT-2 with fine-tuned models on three downstream applications: question answering \cite{wang2022tmlr}, clinical trial matching, and sentence completion. We report accuracy for question answering, NDCG/Precision/Recall for clinical trial matching, and conduct human evaluation for sentence completion. The experimental results show that the fine-tuned Temporal BERT model achieves up to 10\% improvement in terms of accuracy in question anwering, 3\% improvement in clinical trial matching. The fine-tuned GPT-2 reports more clinically reliable answer compared with vanilla GPT-2. 
	
	
Our contributions are summarized as follows:
\begin{itemize}
    \item We release a large scale text time series datasets from 267,284 discharge summaries with 22,588,586 clinical event, timestamp pairs.
    \item We release the Transformer encoder, BERT model fine-tuned on our dataset.
    \item We release the large Transformer decoder, GPT-2 model fine-tuned on our dataset.
    \item This work proposes to use context Retrieval Augmented Generation (RAG) to boost diversity and consistency of content annotation by LLM.
    \item We propose the high-quality prompt strategy for long-term reasoning for clinical event and timestamp annotation.
    \item A novel end-to-end framework is proposed to extract textual time series clinical events from free text clinical notes. 
    \item We design the fine-tuning strategy to validate that the model fine-tuned on our dataset achieves improvement in real-world healthcare tasks, such as medical question answering, clinical trail matching and sentence completion.
\end{itemize}

\section{Related Work}

There are many works to extract clinical events (clinical, epidemiological, demographics) from free-text clinical data which act as important features for treatment recommendation \cite{wang2018provable}. For example, the popular Named entity recognition (NER) techniques, which identifies a named entity (a real-world object or concept) in unstructured text and then classifying the entity into a standard category, can be directly used for clinical event extraction \cite{nadeau2007survey,yadav2018survey,zhong2021frustratingly}. There are some works proposed to recognized biomedical and chemical name entity from the free text data (medical records, clinical notes, case reports, scientific publications) \cite{raza2022large}. However, the state-of-the-art work in biomedical named entity recognition mainly focuses on limited named entities (disease, chemicals, genes, etc.). Our work aims at recognizing anything in the clinical notes related to clinical diagnoses and patients' information, such as disease, symptoms, medical concepts; risk factors, epidemiological entities, such as infectious diseases or patient demographics; patient's information, such as age, sex, admitted to hospital. The goal is to facilitate medical practitioners, clinicians, nurses, and doctors in the fast retrieval of information with high accuracy and efficiency.

\noindent{\bf Temporal-information extraction}. Temporal common sense such as duration and frequency of events is crucial for understanding natural language. However, temporal information extraction is a challenging task because such information is often not expressed explicitly in text. For example, here is a paragraph from case report 	``PMC4478313'':

\begin{figure}[ht]
	\centering
	\fbox{%
		\begin{minipage}{0.9\textwidth}
		``After obtaining the written consent for the off-label use of intravitreal ranibizumab, the patient was scheduled for 3 monthly doses (0.5 mg in 0.05 ml) in his OS. After the third injection, his BCVA improved to 3/10''.
		\end{minipage}
	}
	\label{fg:example}
\end{figure}

 If the off-label use of intravitreal ranibizumab is set at time 0, then the ``third injection'' and the related symptoms ``his BCVA improved to 3/10'' should happen around 2 months later which is not presented in the report. Temporal information is with critical importance for clinical applications. There are some works proposed to extract temporal relationship between events such as  extraction of temporal expressions \cite{lee2014context}, temporal relation extraction  \cite{ning2017structured,vashishtha2019fine}, and timeline construction \cite{leeuwenberg2018temporal}. There are some approaches to identify temporal common sense, such as event duration \cite{zhou2020temporal}, typical temporal ordering \cite{chklovski2004verbocean}, and script learning (i.e., what happens next after certain events) \cite{granroth2016happens,li2018constructing}. There are some multi-task learning that builds two separate models for named entity extraction and the relation extraction \cite{zhong2021frustratingly,wadden2019entity, pustejovsky2003timebank}. In this work, we annotate the relative timestamps of events by setting the admission to hospital as time 0. The use of relative timestamps makes the information easier to interpret and provides a valuable reference for physicians. This is the first dataset to include detailed timestamps for patient-related events. 

 \noindent{\bf Large Language Models for annotation}. In most recently, LLM such as GPT, Claude, Gemini, and Llama, has been widely used to directly generate clinical events from free-text notes with retrieval techniques \cite{jin2024matching,holste2024harnessing,yang2024ensuring,wang2022fast}. However, naively using LLMs, such as ad-hoc prompting with ChatGPT, is far from sufficient for medical tasks for the following reasons. First, there is a lack of high-quality prompts grounded in medical expertise, which limits the reliability of generated outputs. Additionally, challenges arise from the generative nature of LLMs, which often produce random or unreliable content. In practice, identifying clinical events is more akin to a feature selection problem, where key information should be directly extracted from the original clinical notes. Meanwhile, extracting temporal information requires careful reasoning and contextual understanding. There are some methods proposed to draw randomness from uniform distribution \cite{bouras2024integrating}. In this work, we propose to solve the challenge by using context retrieval-augmented generation to provide a candidate pool for LLM for information extraction. To achieve the balance between randomness and diversity, first we provide the candidates of clinical events for LLM to review and select the real clinical events; secondly, we work with clinical expert to propose a prompt strategy that provide detailed examples and reasoning steps to guide LLM for our annotation task. Our framework will be introduced in the following section.

\section{Method}

\noindent{An End-to-End Clinical Event and Temporal Information Annotation Framework}
It consists of two components: contextual retrieval using BM25 and semantic similarity, and LLM based verification and reasoning.

%

\subsection{Contextual Retrieval}
\label{sub:rag}

LLMs tend to generate random and unreliable results. For example, we use Llama to annotate clinical events from the discharge summary, it outputs events like ``Male'' while the summary specifies that the patient is female with ``Sex: F''. To prevent such error, we propose to provide a clinical event candidate dataset which consists of chunks directly retrieved from the discharge summary  Then LLM identifies clinical events and related timestamp with the prompt strategy mentioned in the previous section. The dataset of clinical event candidates are retrieved by contextual BM25 and semantic search with document summary as query. Here, we introduce each component of the retrieval framework. 

\textbf{Dataset}. The original data is the discharge summary table of MIMIC-IV-Note. The discharge summary table contains a ``note$\_$id'' which uniquely identifies a note and discharge summaries for hospitalizations. The summaries are long form narratives which describe the reason for a patient's admission, their hospital course, and any relevant discharge instructions. We use the same number of notes as MIMIC-IV-Ext-BHC dataset, since the summary in the dataset is used as query in our pipeline. All inclusion criteria for MIMIC-IV also apply to this dataset. That is, MIMIC-IV-Ext-BHC selects notes with the brief hospital course section containing more than 100 characters and occurring within one year of a patient encounter (an emergency department visit or a hospital stay). This results in a total of 267,284 clinical notes.

\textbf{Query}. MIMIC-IV-Ext-BHC provides a concise introduction of the patient's hospital course with average token lengths 564. It uses a series of data processing such as whitespace removal, section identificaition, tokenization, and extraneous symbol removal based on the brief hospital course (BHC) section to get a standardized and structured format of the summary. BHC section describes the patient's progress and key interventions, offering a snapshot of the hospitalization. Hence, we use  MIMIC-IV-Ext-BHC as the query to retrieve clinical events related to the patient.

\textbf{Original chunks}. Each discharge summary is segmented into a series of chunks, each containing at most five tokens. The choice of five tokens results in approximately 100 to 400 chunks per summary, which is a manageable size for large language models (LLMs) to process for review and annotation.

\textbf{Context of chunks}. We treat the previous 10 tokens and following 10 tokens of the current chunk as the context. Here is the example of original chunk and related context:

\begin{figure}[ht]
	\centering
	\fbox{%
		\begin{minipage}{0.9\textwidth}
			\textbf{Original Chunk}: \\ 
			\hspace*{1.5em}``clear to auscultation bilaterally, no''. \\[0.5em]  
			
			\textbf{Contextualized Chunk}: \\ 
			\hspace*{1.5em}``Neck: cervical lymphadenopathy. supple, JVP not elevated. \\
			\hspace*{1.5em}Lungs: \textbf{clear} to auscultation bilaterally, no wheezes, rales, rhonchi. \\
			\hspace*{1.5em}CV: regular rate and rhythm.''
		\end{minipage}
	}
	\caption{An example of the original and contextualized chunk demonstrating clinical findings related to lung examination.}
	\label{fg:contextchunk}
\end{figure}
\vspace{-0.2in}
As the example shown in Figure \ref{fg:contextchunk}, the contextualized chunk provides a comprehensive examination report, including findings from multiple systems: such as neck, JVP (Jugular Venous Pressure), lungs, Cardiovascular. The inclusion of related findings supports the exclusion of other potential etiologies of respiratory symptoms, including heart failure and systemic infection.

\textbf{Contextual retrieval methods}. We use two retrieval methods, contextual BM25 and semantic search. The main contribution of the contextual methods is to prepend chunk-specific explanatory context to the original chunk before retrieval and embedding. We incorporate one preceding chunk and one succeeding chunk as context, then combine these with the original chunk for search and embedding.  Here is the steps:
\begin{itemize}
	\item Contextual BM25(best matching): is a ranking function that uses lexical matching to find precise word or phrase matches. It is effective for querying unique identifiers or special terms. BM25 works by building upon the TF-IDF (Term Frequency-Inverse Document Frequency) concept. TF-IDF measures how important a word is to a document in a collection. BM25 refines this by considering document length and applying a saturation function to term frequency. To this end, it helps prevent common words from dominating the results. In our work, the BHC summary is treated as the query, while the original chunks, along with their related context, serve as the candidate pool. We apply BM25 to retrieve the top 100 original chunks most relevant to the query.
	\item Contextual semantic search: we use ``BAAI/bge-large-en'' model \cite{bge_embedding} for embedding which interprets the user’s query for intent, sentiment, and contextual cues. The predefined model ``bge-large-en'' convert the queryand contextual chunks into a 1024-dimensional vector, and their similarity score is computed. We choose 0.75 as the threshold to select the chunks with high similarity with the query. 0.75 is chose as the threshold because it is commonly believed that the text sentences with semantic score higher than 0.75 is believed to be with relatively high similarity. 
\end{itemize}

\textbf{Retrieval process}. Given the brief hospital courses as the query and the contextual chunks, BM25 retrieve 
its nearest neighbors from the pool of candidates, e.g. chunks from discharge summary which are achieved by breaking up the summary in chunks with at least 5 tokens with no gap. The retrieved nearest neighbors are chunks similar to the summary. In the same time, we learn the embeddings of query and contextual chunks, select chunks with a correlation score higher than 0.75.

\begin{table}[ht!]
	\centering
	\caption{Prompt Strategy for Clinical Time Series Event Extraction}
	\begin{tabular}{@{}p{0.28\textwidth}p{0.67\textwidth}@{}}
		\toprule
		\rowcolor{gray!10} \textbf{Category}       & \textbf{Details} \\ \midrule
		\textbf{Task:}     & Extract clinical events and estimate related timestamps  The discharge summary and a list of chunks may contain clinical events. \\ \hline
		\textbf{Event Guidance:} & 1. Identify clinical events **only** from the provided chunks. A clinical event is a free-text specification of an entity pertaining to or with the potential to affect the person's health that can be temporally located.  \\ 
		&2. Include as many clinical events as possible. Each event must come directly from the chunks provided. Do not create or infer events from the full document if they are not explicitly present in the chunks.  \\ 
		&3. For each identified clinical event:  \\ 
		& \hspace{0.5cm} - 3.1 Extract the event directly from the chunk or use it as-is if it is concise enough.  \\ 
		&\hspace{0.5cm} - 3.2 Separate conjunctive phrases into their component events and assign them the same timestamp. For example:``fever and rash'' → ``fever'' | -72 and ``rash'' | -72.  \\
		&	\hspace{0.5cm} - 3.3 Include events with durations (e.g., treatments or symptoms) and assign the time as the **start** of the interval. \\ \hline
		\textbf{Timestamp Guidance:}         & 1. The admission event is always assigned a timestamp of 0.  \\
		& 2. Events occurring before admission have negative timestamps, while vents occurring after admission have positive timestamps, measured in hours.\\
		& 3. Estimate the relative timing of the event:  \\ 
		&\hspace{0.5cm} - 3.1 Base the timing on the **context of the entire document** but ensure the event itself comes from the chunks.  \\ 
		&	\hspace{0.5cm} - 3.2 Use explicit or inferred temporal information to assign a numeric timestamp.  \\ 
		& When explicit timing is not available:  \\ 
		&\hspace{0.5cm} - 3.3 Use inferred durations from the document's context.  \\ 
		&\hspace{0.5cm} - 3.4 Apply clinical judgment to estimate a reasonable time.  \\
		& \hspace{0.5cm} - 3.5 Provide approximate values (e.g., ``a few weeks ag'' → ``-336'' hours).   \\ \hline
		\textbf{Important Notes:} 
		&	1. Only use events that appear in the chunks provided.  \\ 
		&	2. Do not infer or create events from the document if they are not present in the chunks.  \\ 
		&	3. Maximize inclusion of events from the chunks. \\
		\bottomrule
	\end{tabular}
		\label{tb:prompt}
\end{table}

 \section{LLM based Annotation}

We use Llama-3.1-8B as the annotator. The task is to extract time series clinical event and related timestamp. To this end, we provide the definition of a clinical event in Definition \ref{df}. Though there are existing name entity extraction tools, such as cTakes \cite{savova2010mayo}, National Library of Medicine MetaMap \cite{metamap}, PhenoTagger, these tools can retrieve concepts such as symptoms, procedures, diagnoses, medications and anatomy which share similarity with the clinical events in our definition. However, these systems usually depend on a predefined health or biomedical vocabularies and standards, such as the Unified Medical Language System (UMLS) Metathesaurus concepts \cite{aronson2000nlm}. New concepts have emerged that are not yet covered by existing systems. Hence, these systems tend to miss important information since electronic healthcare system generates numerous clinical notes every year. The semanticnetwork in UMLS only covers semantic relationship while most clinical events are conceptually linked but with not semantic correlated. For example, the medicine ``Amiodarone'' is deeply tied to ``cardiovascular disease'' in a clinical context. Because ``Amiodaorone'' is frequently chosen for patients with severe arrhythmias. However,``Amiodarone'' and ``cardiovascular disease''  are not interchangeable or semantically equivalent.

\textbf{Clinical event annotation guidance}. Our framework breaks the barrier of predefined dictionary and guides LLM to select all related terms related to the patient. Here is the brief guidance for clinical event identification:
\begin{itemize}
	\item Identify clinical events only from the provided chunks, which are retrieved by contextual BM25 and semantic search.
	\item Include as many events as possible.
	\item Separate conjunctive phrases into their component events, for example, convert ``fever and rash'' to two events ``fever'' and ``rash''.
\end{itemize}

The goal to estimate timestamp of clinical event is to build the medical history of the patient. It is fundamental to applications such as understanding treatment progression, assessing outcomes, and informing future clinical decisions. The temporal information, especially onset and duration is critical for physicians to make treatment decision. The most popular temporal related question asked by physician after the symptom description is 
\begin{example}
\textbf{Temporal Questions}: ``When did the symptoms first start?'', ``How long do the symptoms last each time they occur?''
\end{example}
To this end, we ask LLM to identify a pivotal clinical event and designate its timestamp as zero, usually the admission event. From there, LLM is trained to measure the timing of all other events relative to this pivot event, creating a standardized timestamp that simplifies clinical interpretation and decision-making. 

\textbf{Clinical Temporal Information Annotation guidance}. Here is the timestamp annotation guidance:
\begin{itemize}
	\item Estimate the timestamp of events based on the entire discharge summary;
	\item Use explicit or inferred temporal information to assign a timestamp;
	\item Apply clinical judgment to estimate a reasonable time when explicit timing is not available;
	\item Format timestamp in hours. 
\end{itemize} 

The detailed prompt strategy is presented in Table \ref{tb:prompt}. The input of LLM other than the prompt are shown in Table \ref{tb:input}, which includes the chunks of discharge summary which is a dataset of clinical event candidates.
 
%
%

\begin{table}[ht]
	\centering
	\caption{Clinical Time Series Event Extraction: Input and Output Details}
\begin{tabular}{@{}p{0.2\textwidth}p{0.75\textwidth}@{}}
		\toprule
		\rowcolor{gray!10} \textbf{Category}       & \textbf{Details} \\ \midrule
		\textbf{Input:}  &  \\ 
		\textbf{Document:}     & A comprehensive discharge summary providing the patient’s history and context. \\ 
		\textbf{Chunks:}         & A list of small text segments (3 to 10 tokens each) extracted from the document. \\ \hline
		\textbf{Output:}  &  \\ 
		\textbf{Text Time Series:}    & The events and related timestamps in a list. Each item has two elements: the clinical event and the estimated relative timestamp in hours, separated by a pipe ($|$). \\ 
		\textbf{Example:} & \texttt{fever | -72} \\ \bottomrule
	\end{tabular}
	\label{tb:input}
\end{table}

 \noindent{\textbf{Post processing for LLM based Annotation}}. There are errors and inconsistencies in the output of Llama-3.1-8B annotation. For example, it sometimes produces invalid timestamp annotations such as ``NaN'' or ``in'', duplicates clinical events, misplaces information by writing clinical events under the ``Time'' column and timestamps under the ``Event'' column; includes non-textual entries as clinical events; and repeatedly outputs phrases like “no acute process.”

 \noindent{\textbf{Validation of Prompt Strategy}}. We validate the effectiveness of our prompting strategy on pulic dataset case reports by comparing with GPT-4, O1-preview and clinical expert annotation. There are 10 randomly selected case reports and annotated by a medical doctor: PMC4478313, PMC4818304, PMC5667582,
 PMC6030904, PMC6034490, PMC7337692, PMC7747049, PMC8127753, and PMC9871993. Table \ref{tab:annote} shows an example of annotation for case report PMC4300884. 
 
 \begin{table}[]
 	\caption{The example of annotated events and timestamps on case report PMC4300884.}
 	\begin{tabular}{|l|l|}
 		\hline
 		Events                                                                              & Timestamps \\ \hline
 		Onset of right-sided L5 radiculopathy symptoms                                      & -4320      \\ \hline
 		Presentation to clinic with symptoms                                                & -1440      \\ \hline
 		Imaging studies (X-ray and CT scan) showed large anterior osteophyte at L5-S1       & -1440      \\ \hline
 		Selective transforaminal L5-S1 nerve root injection performed                       & -720       \\ \hline
 		Temporary improvement of symptoms after nerve root injection                        & -720       \\ \hline
 		Nonoperative therapies exhausted; patient elected surgical intervention             & -168       \\ \hline
 		Surgery performed: L5-S1 anterior lumbar interbody fusion with osteophyte resection & 0          \\ \hline
 		Immediate postoperative resolution of leg pain                                      & 0          \\ \hline
 		Six months after surgery  patient remains symptom free and in rehabilitation        & 4320       \\ \hline
 	\end{tabular}
 	\label{tab:annote}
 \end{table}

 GPT-4 and O1-preview use the prompt as defined in the previous section.  Table \ref{tab:compare} shows the average number of annotations for all three methods. For example,the number of events annotated by clinical expert ranges from 14 to 70, GPT-4 reports 20 to 297 events while O1-preview selects 27 to 58 events. There are on average 6 distint timestamps for all three methods. While the manual annotation results in fewer events as the LLM annotations (Figure 2, left), the event match rate at
 a cosine distance threshold from 0.1 ranges between 70–80\%. Comparing the relative times of matched events with that of the manual annotations, we find that the LLM
 annotations possessed high concordance (means—GPT-4: 0.912, and O1-preview: 0.951). It validae the effectiveness of our prompt. The detailed results are shown in \cite{summit25}.
\begin{table}[]
		\caption{The difference between manual annotation and LLM with our prompt.}
	\begin{tabular}{|l|l|l|l|}
		\hline
		Statistics & Manual         & GPT-4          & O1-preview    \\ \hline
		Events     & 32 {[}14,70{]} & 46{[}20,297{]} & 39{[}27,58{]} \\ \hline
		Timestamps & 6{[}2,13{]}    & 6{[}1,16{]}    & 6{[}3,13{]}   \\ \hline
	\end{tabular}
		\label{tab:compare}
\end{table}
We choose Llama-3.1-8B as our annotator for the comprehensive consideration of restriction of MIMIC dataset,  computation efficiency and annotation quality.

 
 
  \begin{figure}[htp]
 	\centering
 	\begin{subfigure}[b]{0.75\textwidth}
 		\centering
 		\includegraphics[width=0.8\textwidth]{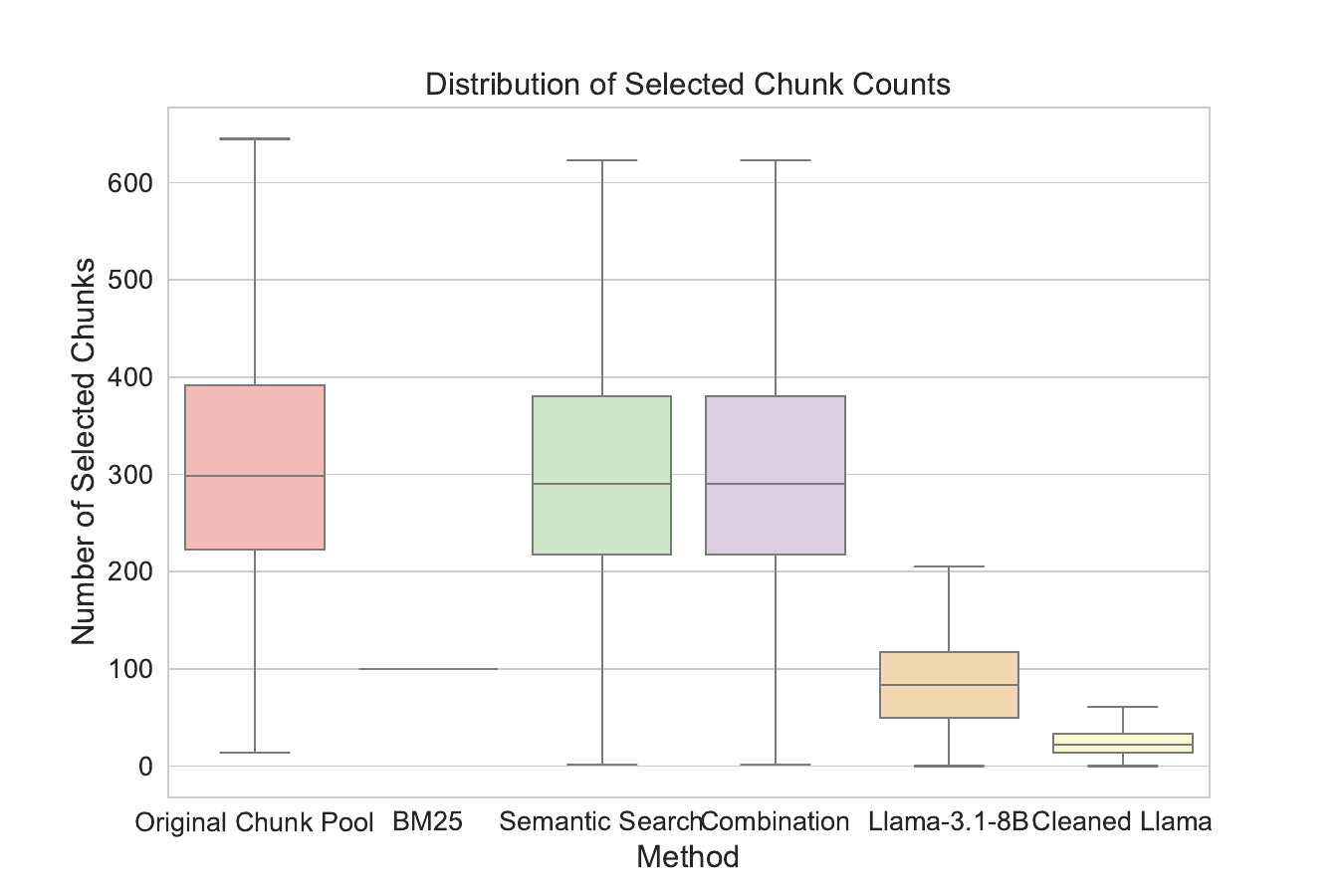} 
 		\caption{The distribution of original chunk pool and selected chunks by each step. For example, there are 200 to 400 chunks for each discharge summary in the original chunk pool. Contextual BM25 selects 100 chunks per summary. Contextual semantic search also selected 200 to 400 chunks per summary, similarly for the combination (BM25 + semantic search). Llama-3.1-8B selects 50 to 100 chunks per summary, after data cleaning and normalization, there are less than 50 chunks per summary.}
 		\label{fig:numb}
 	\end{subfigure}
 	
 	\vspace{0.2in} 
 	
 	\begin{subfigure}[b]{\textwidth}
 		\centering
 		\includegraphics[width=0.45\textwidth]{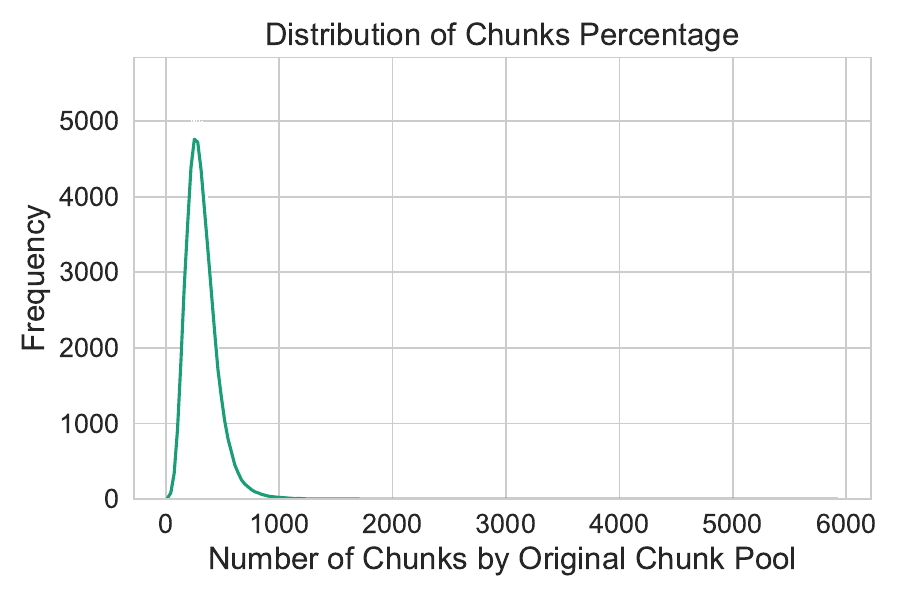}
 		\includegraphics[width=0.45\textwidth]{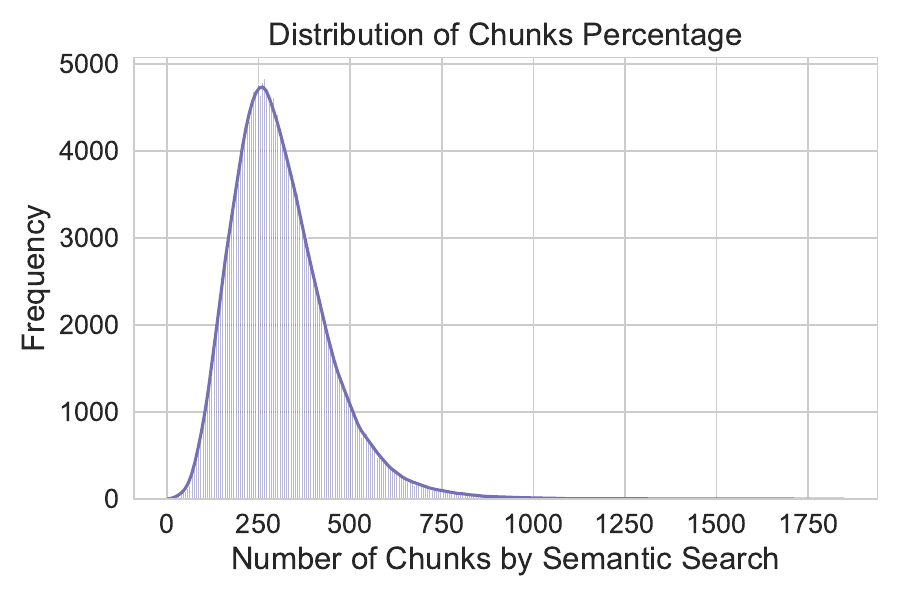}
 		
 		\includegraphics[width=0.45\textwidth]{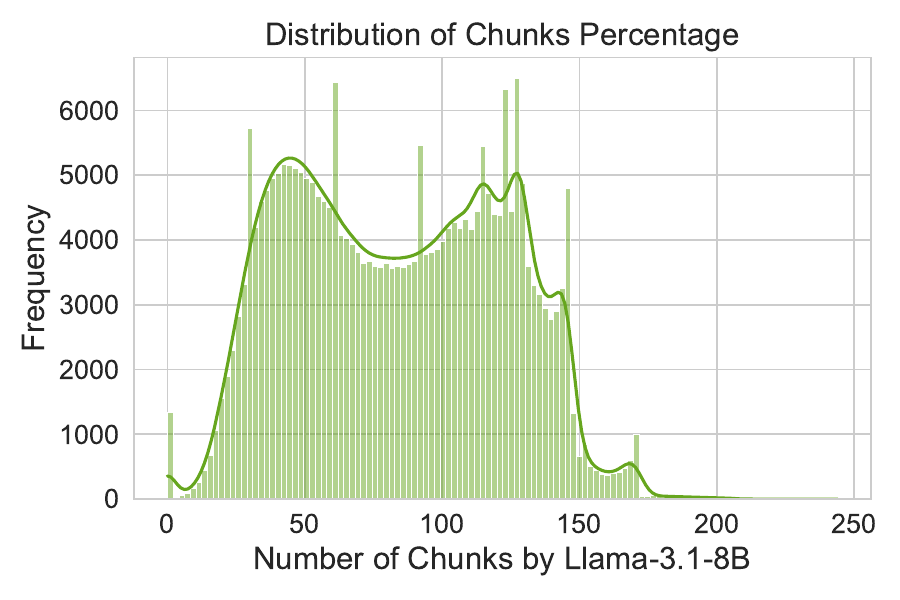}
 		\includegraphics[width=0.45\textwidth]{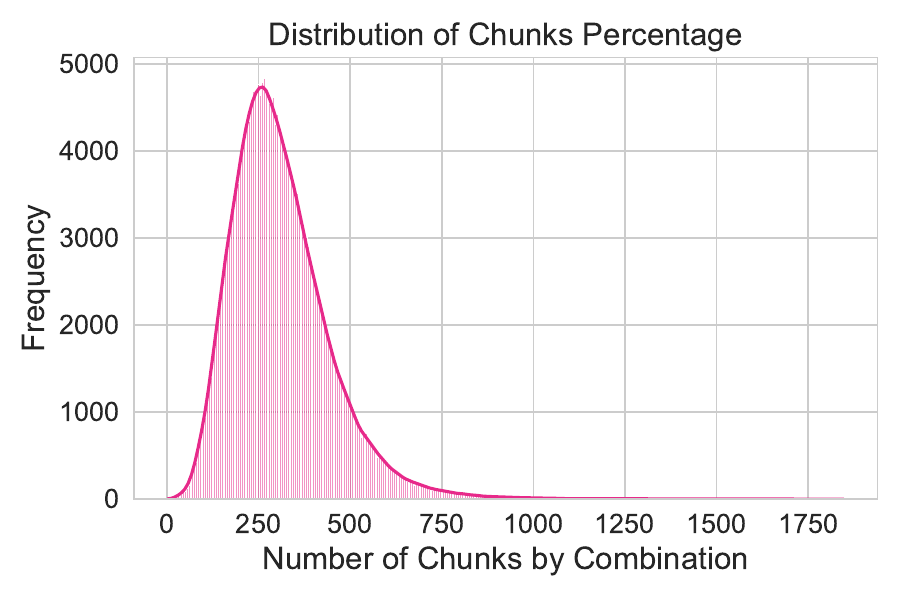}
 		
 		\includegraphics[width=0.7\textwidth]{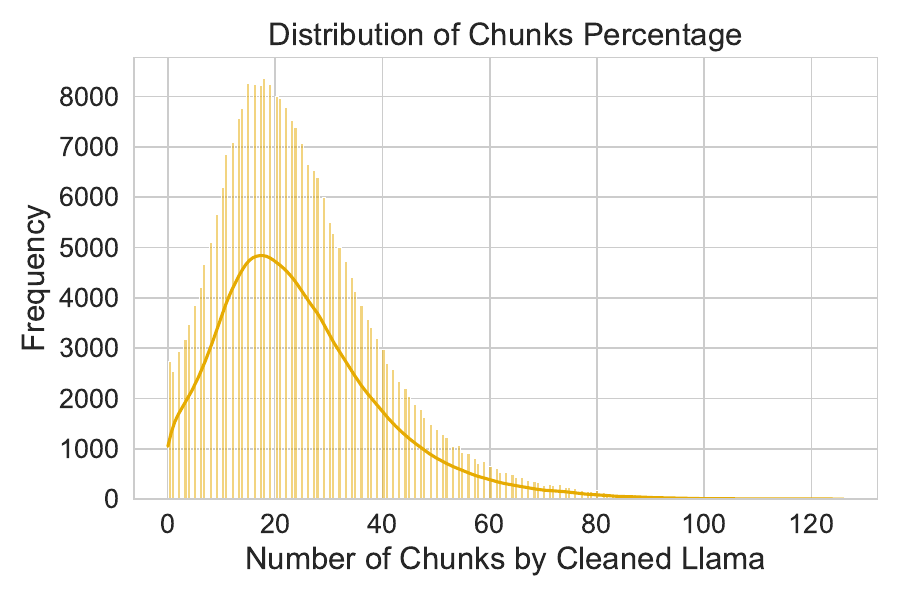}
 		
 		\caption{The histogram of chunk frequency at each step. For example, contextual BM25 selects around 30\% of the original chunks.}
 		\label{fig:each}
 	\end{subfigure}
 	
 	\caption{Comparison of chunk selection distribution and histogram of chunk frequency at each step.}
 \end{figure}
 
 \section{Result}
 The dataset consists of 22,588,586 clinical events and related timestamps from 267,284 discharge summaries from MIMIC-IV-Note dataset. There is at least 1 clinical event and timestamp for each summary. The maximal number of clinical event and timestamp is 244. The average number of clinical event and timestamp is 84. The average number of tokens per clinical event is 3. There are at most 299 tokens per clinical event. The timestamp of clinical event is negative if the event happens before admitted to the hospital, and positive if happens after admitted to the hospital. The timestamp is in hours. There are 36.99\% of historical clinical events, 51.19\% of clinical events during admission, and 11.80\% events happen after admission. 
 
 There are five columns in the table: \textit{Hadm\_id}, \textit{Event}, \textit{Time}, and \textit{Time\_bin}. The column \textit{Hadm\_id} serves as a unique identifier for each discharge summary, ensuring that each discharge note (MIMIC-IV-Note) and its associated query (MIMIC-IV-Ext-BHC) can easily be referenced. The column Event shows the clinical event in text format, the column Time shows the timestamp that the event happens. \textit{Time\_bin} is obtained by mapping the continuous temporal annotations into discrete categories.  The timestamp is assigned to one of the following predefined intervals: $[-\infty, -60, -30, -15, 0, 15, 30, 60, 120, \infty]$. The temporal annotation is placed into exactly one bin according to these boundaries.
 
  For each step in the pipeline, there are around 200 to 400 chunks for each discharge summary after breaking the discharge summary into chunks with 5 tokens. The contextual BM25 selects 100 chunks per summary. The semantic search retrieves 200 to 400 chunks per summary with correlation threshold 0.75. The combined retrieval result is around 200 to 400 chunks per summary.  Llama-3.1-8B identifies 50 to 100 chunks that contains clinical events based on the retrieval result. 
  
  We show the distribution of selected chunks compared to the original number of chunks as shown in Figure \ref{fig:numb}. 
Figure \ref{fig:each} shows the chunks frequency for each step. For example, there are around 8,000 discharge summary has 20 chunks after Llama-3.1-8B annotation.
\subsection{Fine-tuning BERT}


We fine-tuned BERT for a correlation classification problem, where the correlation between two events is labeled as: no correlation (0), positive correlation (1), or negative correlation (2). Given the events $E_a$ and $E_b$, from the same patient, where $E_a$ appears before $E_b$ in the case report. Their correlation is based on their temporal annotation. If the annotated timestamps are the same ($T_a=T_b$), the events are considered simultaneously and labeled as having no correlation (0). If $E_b$ occurs after $E_a$ ($T_b>T_a$), we assign a positive correlation label (1), indicating that $E_b$ is the outcome or consequence of $E_a$. Conversely, if $T_b < T_a$, the correlation is labeled as negative (2), implying that $E_b$ is a possible cause or antecedent of $E_a$. 

\noindent{\textbf{Data processing for BERT}}. We transform the continuous temporal annotations into discrete categories by assigning each annotation to one of the following predefined intervals: $[-\infty, -60, -30, -15, 0, 15, 30, 60, 120, \infty]$. 
Each temporal annotation is placed into exactly one bin according to these boundaries. The final dataset is a sequence of data pair in the format of $\{\langle E_a, E_b, y, t \rangle \}_n$, where $E_a$ and $E_b$ are textual clinical event, $t\in [0,9]$ is index of the bins, $n=22,321,430$ is the final dataset is a sequence of data pairs, $y\in [0,2]$ is the correlation label. Here are some examples of the data paris, $\langle$laxative-induced diarrhea, pain control, 0, 4$\rangle$, $\langle$choline 
\& magnesium salicylate 500 mg/5 mL Liquid, atenolol 50 mg Tablet, 0, 4$\rangle$. 

We randomly select 80\% samples for training  20\% for validation. We fine-tune BERT model using HuggingFace ``bert-base-uncased''. 

\textbf{Model design.} To effectively handle both textual and temporal information, we design the model to jointly incorporate these modalities for downstream tasks. It consists of the following components:
\begin{itemize}
	\item Text encoding: the default BERT model that maps the clinical events into a $h_{CLS}\in R^H$ vector, as
	\begin{align}
		h_{CLS} = \text{BERT}(\text{input}, \text{attention}\_\text{mask})
	\end{align}
	\item Time embedding: the embedidng layer maps the time bin to a $d$-dimensional dense vector. 
\begin{align}
	h_{\text{time}} = \mathrm{Embedding}(\text{time\_bin}).
\end{align}
	\item Feature fusion: concatenate the BERT-based represetantion with the tim embedding:
 \begin{equation}
	z = [\,h_{\text{CLS}};\,h_{\text{time}}\,] \in \mathbb{R}^{H + d}.
\end{equation}
	\item Classifier: a Multi-Layer Perceptron (MLP) that outputs class logits, consisting of a linear layer that reduces dimension and combines features, a ReLU activation for non-linearity, and a dropout layer to prevent overfitting:
\begin{align*}
	\mathbf{z}_1 &= W_1 \, \mathbf{z} + \mathbf{b}_1, ~~
	\mathbf{z}_2 = \mathrm{ReLU}\bigl(\mathbf{z}_1\bigr), \\
	\mathbf{z}_2' &= \mathrm{Dropout}\bigl(\mathbf{z}_2\bigr), ~~
	\mathrm{l} = W_2 \, \mathbf{z}_2' + \mathbf{b}_2.
\end{align*}
where $W_1$, $b_1$, $W_2$ and $b_2$ are the weigth matrices and biases of the two linear layers, $\mathrm{l}$ is the final output which has the dimensionality equal to the number of classes of the classification class.
\end{itemize}

We employ cross-entropy as the loss function. The empirical risk of the classifier $f$ is defined as:
\begin{align}
	R_L(f) = E_D[L(f(x;\theta),y_x)]=-\frac{1}{n}\sum_{i=1}^{n}\sum_{j=1}^c y_{ij} \log f_j(x_n, \theta),
\end{align} 
where $f$ is Temporal BERT model that maps the input feature space to the label space $f: X->R^c$, $\theta$ is the set of parameters of the classifier. The training process is to minimize the risk by optimizing the model parameters. We use the Adam optimizer with a learning rate of $2\times 10^{-5}$. To enable mixed precision training and conserve memory, we incorporate a gradient scaler. We train for a total of five epochs.

\subsection{Question Answering}

We compare the performance of BERT and fine-tuned temporal BERT on PubMedQA dataset \cite{jin2019pubmedqa}. It is a biomedical question answering dataset collected from PubMed abstracts. We use its 1,000  expert annotated subset which consists of question, context, and a yes/no/maybe answer. There are 10 folds of training dataset, each of which has 450 samples. The testing dataset is made of 500 samples. We fine-tune the BERT model and our Temporal BERT model with the same classifier head with the input dimension 768, hidden dimension 256. The optimizer is Adam with learning rate 3e-5. We fine-tune BERT and Temporal BERT for 3 epochs with batch size 16. We report the accuracy, defined as 
\begin{align}
	\textit{Accuracy}(y,\hat{y}) =\frac{1}{n}\sum_{i=0}^{n-1} I(\hat{y}_i=y)
\end{align}
where $n$ is the number of testing samples, $\hat{y}$ is the predicted label. The results are shown in Table \ref{tb:pubmed}. 

\begin{table}[]
	\caption{The comparison of accuracy of Bert and Temporal Bert on PubMedQA dataset}
	\label{tb:pubmed}
	\begin{tabular}{|l|ll|}
		\hline
		Fold & \multicolumn{2}{c|}{Accuracy}              \\ \hline
		& \multicolumn{1}{l|}{Bert}  & Temporal Bert \\ \hline
		0    & \multicolumn{1}{l|}{53.6}  & 56.2          \\ \hline
		1    & \multicolumn{1}{l|}{50.00} & 56.66         \\ \hline
		2    & \multicolumn{1}{l|}{46.00} & 55.2          \\ \hline
		3    & \multicolumn{1}{l|}{47.00} & 55.2          \\ \hline
		4    & \multicolumn{1}{l|}{50.00} & 55.2          \\ \hline
		5    & \multicolumn{1}{l|}{47.6}  & 55.2          \\ \hline
		6    & \multicolumn{1}{l|}{44.2}  & 42.6          \\ \hline
		7    & \multicolumn{1}{l|}{47.00} & 55.2          \\ \hline
		8    & \multicolumn{1}{l|}{48.4}  & 55.2          \\ \hline
		9    & \multicolumn{1}{l|}{44.00} & 55.2          \\ \hline
		mean/variance & \multicolumn{1}{l|}{47.77$\pm$7.57} & \multicolumn{1}{l|}{54.07$\pm$16.72} \\ \hline
	\end{tabular}
\end{table}

\subsection{Clinical Trial Matching}

Clinical trials are experiments done in the development of new treatments, drugs or medical devices. It is expensive and time cost for drug company to recruit patient. Though the clinical trails are publicly available on ClinicalTrials.gov. It is difficult for patient to find a clinical trail to receive experimental treatments that potentially improve the health outcomes. We test the effectiveness of our annotated dataset on the clinical trial retrieval task with a naive model, that computes the correlation between patient and trial directly.

 We fine-tune BERT and Temporal BERT with a test collection for patient-trail matching \cite{koopman2016test}. It has 59 patient case reports used as topics from Trec 2014 and 2015. Each patient case topic has two forms: a description (78 token on average) and a summary (22 tokens on average) which describes a patient with certain conditions and observations. We use the description as the patient information $p_i$. There are 204,855 clinical trails from ClinicalTrials.gov. We use the the clinical trial title, summary and inclusion criteria as the clinical trial information $r_i$. There are 3 classes of correlation between the patient and clinical trial: (1) irrelevant (0), the clinical expert would not refer patient $p_i$ for the clinical trail $r_i$; (2) potential (1), the clinical expert would consider referring patient $p_i$ to the clinical trail $r_i$; (3) eligible (2), the clinical expert would highly likely refer patient $p_i$ to the clinical trail $r_i$. 

\begin{table}[]
	\caption{The statistics of the Training dataset and Testing dataset of Clinical Trial Matching Task.}
	\begin{tabular}{|l|r|r|r|}
		\hline
		& Training Dataset         & TREC 2021     & TREC 2022     \\ \hline
		\#patient                 & 59            & 75            & 50            \\ \hline
		\#trials                  & 204,855       & 375,580       & 448,631       \\ \hline
		age of patient            & 38.5$\pm$23.7 & 41.6$\pm$19.4 & 35.3$\pm$20.2 \\ \hline
		sex (male/female)         & 50/50         & 50.6/49.4     & 56/44         \\ \hline
		irrelevant trials/patient & 46$\pm$379    & 323$\pm$8624  & 568$\pm$26926 \\ \hline
		potential trials/patient  & 11$\pm$103    & 80$\pm$3641   & 60$\pm$4291   \\ \hline
		eligible trials/patient   & 7$\pm$45      & 74$\pm$2403   & 78$\pm$4528   \\ \hline
	\end{tabular}
	\label{tb:trial}
\end{table}
We test the performance of fine-tuned models on TREC 2021 and TREC 2022. There are 3 classes of correlation between the patient and clinical trial: not relevant, excludes (not sufficient information to be qualified), and eligible. The patient topics are created by individuals with medical training. It consists of a synthetic case in the form of an admission note. For example, here is an example of the patient topic in \href{https://www.trec-cds.org/2022.html}{TREC 2022}. There are 75 patient topics in TREC 2021, 50 in TREC 2022. The statistics of the training and testing dataset is shown in Table \ref{tb:trial}.

\begin{figure}[ht]
	\centering
	\fbox{%
		\begin{minipage}{0.9\textwidth}
			\textbf{Embedding Description:} \\
			A 2-year-old boy is brought to the emergency department by his parents for 5 days of high fever and irritability. The physical exam reveals conjunctivitis, strawberry tongue, inflammation of the hands and feet, desquamation of the skin of the fingers and toes, and cervical lymphadenopathy with the smallest node at 1.5 cm. The abdominal exam demonstrates tenderness and enlarged liver. Laboratory tests report elevated alanine aminotransferase, white blood cell count of 17,580/mm, albumin 2.1 g/dL, C-reactive protein 4.5 mg, erythrocyte sedimentation rate 60 mm/h, mild normochromic, normocytic anemia, and leukocytes in urine of 20/mL with no bacteria identified. The echocardiogram shows moderate dilation of the coronary arteries with possible coronary artery aneurysm.
		\end{minipage}
	}
	\caption{A synthetic patient topic in TREC 2022.}
	\label{fig:embedding-description}
\end{figure}

Figure \ref{fig:example-figure} is an example of the clinical trial information with brief title, brief summary and inclusion information that would be used in our experiment. 

\begin{figure}[ht]
	\centering
	\fbox{%
		\begin{minipage}{0.9\textwidth}
			\noindent
			\textbf{Brief title:} Decitabine and Peripheral Stem Cell Transplantation in Treating Patients Who Have Relapsed Following Bone Marrow Transplantation for Leukemia, Myelodysplastic Syndrome, or Chronic Myelogenous Leukemia
			
			\vspace{0.5em} 
			\noindent
			
			\textbf{Brief summary:} RATIONALE: Peripheral stem cell transplantation may be an effective treatment for leukemia;
			myelodysplastic syndrome, or chronic myelogenous leukemia that has relapsed following bone;
			marrow transplantation;
			PURPOSE: Phase I/II trial to study the effectiveness of decitabine and peripheral stem cell;
			transplantation in treating patients who have leukemia, myelodysplastic syndrome, or chronic;
			myelogenous leukemia that has relapsed after bone marrow transplantation;
			
			\vspace{0.5em}
			\noindent
			
			\textbf{Criteria:}
		DISEASE CHARACTERISTICS: Acute leukemia, myelodysplastic syndromes or chronic myelogenous;
		leukemia (CML) in accelerated phase or blast crisis and relapsed within 1 year after;
		allogeneic bone marrow transplantation Must not be candidates for second course of high;
		dose chemoradiotherapy;
		PATIENT CHARACTERISTICS: Age: 60 and under Performance status: Zubrod 0-2 Life expectancy;
		Not specified Hematopoietic: Not specified Hepatic: Bilirubin less than 3 mg/dL Renal;
		Creatinine less than 2 mg/dL Cardiovascular: Greater than 40
		scan or ECHO Other: Not pregnant No serious intercurrent illness No active CNS disease Must;
		be ineligible for protocols of higher priority No active acute graft vs host disease (GVHD);
		greater than grade 2 or extensive chronic GVHD No active uncontrolled infection Original;
		marrow donor must undergo filgrastim primed peripheral blood stem cell collection;
		PRIOR CONCURRENT THERAPY: See Disease Characteristics
		\end{minipage}
	}
	\caption{Example of a clinical trial ``NCI-G96-1000'' with title, summary, and criteria.}
	\label{fig:example-figure}
\end{figure}

We use the retrieval results for each of topic released by
\cite{jin2024matching} as the initial result. Then we compute the correlation score between the topic and trials by BERT and Temporal BERT. Then we report NDCG, Precision and Recall with top 10 and top 100 retrieved results as shown in Table \ref{tb:trec}. The result shows that the model trained on our dataset achieves consistent advantage in all the evaluation criteria.

\begin{table}[ht]
	\centering
	\caption{Evaluation results on TREC 2021 and TREC 2022 of NDCG, Precision, and Recall at cutoffs 10 and 100 (TC21 and TC22 represents TREC 2021 and TREC 2022, T/BERT is Temporal BERT).}
	\label{tb:trec}
	\begin{tabular}{llrrrr}
		\toprule
		\textbf{Evaluation} & \textbf{Cutoff} & \textbf{TC21 BERT} & \textbf{TC21 T/BERT} & \textbf{TC22 BERT} & \textbf{TC22 T/BERT} \\
		\midrule
		NDCG     & @10   & 33.28 & 36.53 & 29.43 & 29.94 \\
		& @100  & 33.19 & 35.15 & 22.69 & 26.86 \\
		\midrule
		Precision & @10   & 49.06 & 50.13 & 34.00 & 36.00 \\
		& @100  & 39.86 & 40.42 & 23.76 & 27.68 \\
		\midrule
		Recall    & @10   & 3.57  & 3.52  & 2.85  & 2.52  \\
		& @100  & 29.01 & 30.39 & 16.01 & 21.57 \\
		\bottomrule
	\end{tabular}
\end{table}

%

\subsection{Fine-tuning GPT-2}

We develop a multi-step pipeline to convert the raw sequence of event and timestamp for each patient into tokenized sequences suitable for GPT-2 modeling. We combine the time series clinical events and timestamps of all discharge summaries, which is a data matrix with two columns ``Event'' and ``Time''. To split patients among training, validation, and test sets, we collect the unique discharge summary ids and shuffle them randomly. We partition the notes into approximately 80\% training, 10\% validation, and 10\% test. At last, we create three subsets with non-overlapping notes. 

We initialize a pretrained GPT-2 tokenizer and augment its vocabulary with two domain-specific tokens: \texttt{[TIME]} and \texttt{[EVENT]}.The dataset is tokenized with a maximum sequence length of 128 tokens, and any empty or zero-length samples are discarded. For training, we set the hyperparameters as follows: batch size of 8, learning rate of 1e-5, and 3 training epoch. To evaluate generation performance, we compare the responses of the original GPT-2 and the fine-tuned GPT-2 by querying a set of medically related questions, as illustrated in Table~\ref{tb:gpt2-comparison}.

\begin{table}[ht]
	\centering
	\caption{Comparison of generated answers by GPT-2 and Fine-tuned GPT-2 on the same question.}
	\label{tb:gpt2-comparison}
	\begin{tabular}{p{5cm} p{3.5cm} p{3.5cm}}
		\hline
		\textbf{Question} & \textbf{GPT-2} & \textbf{w/GPT-2} \\
		\hline
		\textbf{\# 1:} 
		``What are the recommended considerations for prescribing Paxlovid to a 
		patient with diabetes and mild kidney dysfunction?'' 
		& Paxlovid
		& Atorvastatin
		\\
		\hline
		\textbf{\# 2:} 
		``For patients undergoing chemotherapy, what is the usual timestamp for cycles and follow-up imaging to assess treatment response?''
		& The first step is to determine the patient's age and the number of cycles. The second step is to determine the patient's age and the number of follow-up imaging.
		& Cycle 1 of chemotherapy with cisplatin and etoposide.
		\\
		\hline
		\textbf{\# 3:} 
		``A patient underwent major abdominal surgery and is developing tachycardia, 
		hypotension, and reduced urine output in the recovery room. What criteria 
		indicate they should be monitored in the ICU, and what are potential 
		interventions?''
		& The ICU is the place where the patient is most at risk for complications.
		& Patient was given IV fluids until tolerating oral intake, and was 
		given IV pain medications.
		\\ \hline
	\end{tabular}
\end{table}

Given the 1st question, GPT-2 with fine-tuning on our dataset suggests ``Atorvastatin''. ``Atorvastatin'' is long-term therapy to help lower LDL cholesterol, reduce triglycerides (in many cases), and reduces the risk of heart attack and stroke—particularly important in patients with diabetes, who already have a higher cardiovascular risk. ``Statins'' like ``atorvastatin'' do not typically require dose adjustment in mild kidney dysfunction. They can usually be used safely in patients with mild to moderate kidney issues. GPT-2 recommended ``paxlovid'' which is prescribed for a short course (usually 5 days) in eligible patients with COVID‑19 to reduce the risk of severe illness and hospitalization. ``Paxlovid'' doses must be adjusted for moderate kidney dysfunction  (eGFR 30$–$59 mL/min), and it is not recommended (or used cautiously with further dose adjustments) in severe kidney dysfunction (eGFR $<$30 mL/min). 

For the 2nd question, GPT-2 mostly produces irrelevant and repetitive result. It does not mention actual timestamp or imaging schedules. The answer of fine-tuned GPT‑2 is more clinically oriented, closer to a chemotherapy context.

For the 3rd question, the statement of GPT-2 is too broad to be clinically useful.  Fine-tuned GPT‑2 offers more detail by mentioning treatments, IV fluids and IV pain medications. It is an initial management perspective, which could be part of postoperative care.  Thought its statement is not comprehensive, shows more specificity about typical postoperative measures.

\section{Conclusion}

In this work, we introduce the first large-scale time series dataset of clinical events and timestamps, comprising over 22 million examples. This dataset is the first of its kind to incorporate temporal information, making it a valuable resource for advancing healthcare analytics and hospital preparedness for disease outbreaks. The dataset can be noisy, however, the learning algorithm based on the dataset may be inefficient but achieves certain error rate \cite{shen2021sample,shen2019robust}.

Our dataset captures the trajectory of patients with various diseases, enabling the development of predictive models for clinical risk forecasting and causal reasoning. We demonstrate its utility by pretraining a BERT model, achieving up to a 10\% improvement in accuracy for clinical question answering and a 3\% improvement in clinical trial matching. Additionally, a GPT-2 model fine-tuned on our dataset generates more clinically relevant responses.

We believe this dataset will benefit a wide range of deep learning models in healthcare by serving as a foundation for analysis and fine-tuning, ultimately improving clinical decision support and patient outcomes. In the future, we aim to consider fairness in annotation \cite{shen2022metric}, while the current version treat each note independently.

\section{Acknowledge}

This resea­rch was suppo­rted by the Divis­ion of Intra­mural Resea­rch of the Natio­nal Libra­ry of Medic­ine (NLM)­, Natio­nal Insti­tutes of Healt­h. This work utilized the computational resources of the NIH HPC Biowulf cluster \footnote{\href{https://hpc.nih.gov}{Biowulf cluster}}.

\bibliographystyle{plainnat}
\bibliography{snbib}

\end{document}